\documentclass[10pt,twocolumn,letterpaper]{article}

\usepackage{iccv}              

\usepackage{amssymb}  
\usepackage{pifont}   
\usepackage{amssymb}  
\usepackage{pifont}   
\usepackage{colortbl} 
\usepackage{booktabs} 
\usepackage{amsmath}
\usepackage{mathtools}
\usepackage{xcolor}
\usepackage{tikz}
\usepackage{subcaption}
\usetikzlibrary{positioning}
\newcommand{\xmark}{\ding{55}} 
\usepackage{algorithm}
\usepackage{algpseudocode}
\usepackage{setspace}

\definecolor{iccvblue}{rgb}{0.21,0.49,0.74}
\usepackage[pagebackref,breaklinks,colorlinks,allcolors=iccvblue]{hyperref}
\usepackage{graphicx}
\usepackage{amsmath}
\usepackage{amssymb}
\usepackage{colortbl}
\usepackage{graphicx}
\usepackage{xcolor}
\usepackage{multirow}
\usepackage{booktabs} 
\usepackage{amssymb}
\usepackage{pifont}
\usepackage{algorithm}
\usepackage{algpseudocode}

\def\paperID{6882} 
\def\confName{ICCV}
\def\confYear{2025}
\title{Space Rotation with Basis Transformation for \\Training-free Test-Time Adaptation}

\author{{Chenhao Ding\textsuperscript{\rm 1}} \quad
Xinyuan Gao\textsuperscript{\rm 1} \quad 
Songlin Dong\textsuperscript{\rm 2} \quad 
Yuhang He\textsuperscript{\rm 2}   \\
Qiang Wang\textsuperscript{\rm 1} \quad 
Xiang Song \textsuperscript{\rm 1} \quad
Alex Kot \textsuperscript{\rm 3} \quad 
Yihong Gong\textsuperscript{\rm 1,} \textsuperscript{\rm 2}\quad 
\vspace{0.2em} \\
\textsuperscript{\rm 1}School of Software Engineering, Xi’an Jiaotong University \\
\textsuperscript{\rm 2}College of Artificial Intelligence, Xi’an Jiaotong University \\
\textsuperscript{\rm 3}Nanyang Technological University \\
}

\begin{document}
\maketitle
\begin{abstract}
With the development of visual-language models~(VLM) in downstream task applications, test-time adaptation methods based on VLM have attracted increasing attention for their ability to address changes distribution in test-time. Although prior approaches have achieved some progress, they typically either demand substantial computational resources or are constrained by the limitations of the original feature space, rendering them less effective for test-time adaptation tasks. To address these challenges, we propose a training-free feature space rotation with basis transformation for test-time adaptation. By leveraging the inherent distinctions among classes, we reconstruct the original feature space and map it to a new representation, thereby enhancing the clarity of class differences and providing more effective guidance for the model during testing. Additionally, to better capture relevant information from various classes, we maintain a dynamic queue to store representative samples. Experimental results across multiple benchmarks demonstrate that our method outperforms state-of-the-art techniques in terms of both performance and efficiency.
\end{abstract}    
\section{Introduction}
\label{sec:intro}

\begin{figure*}[ht]
\centerline{\includegraphics[width=1\linewidth]{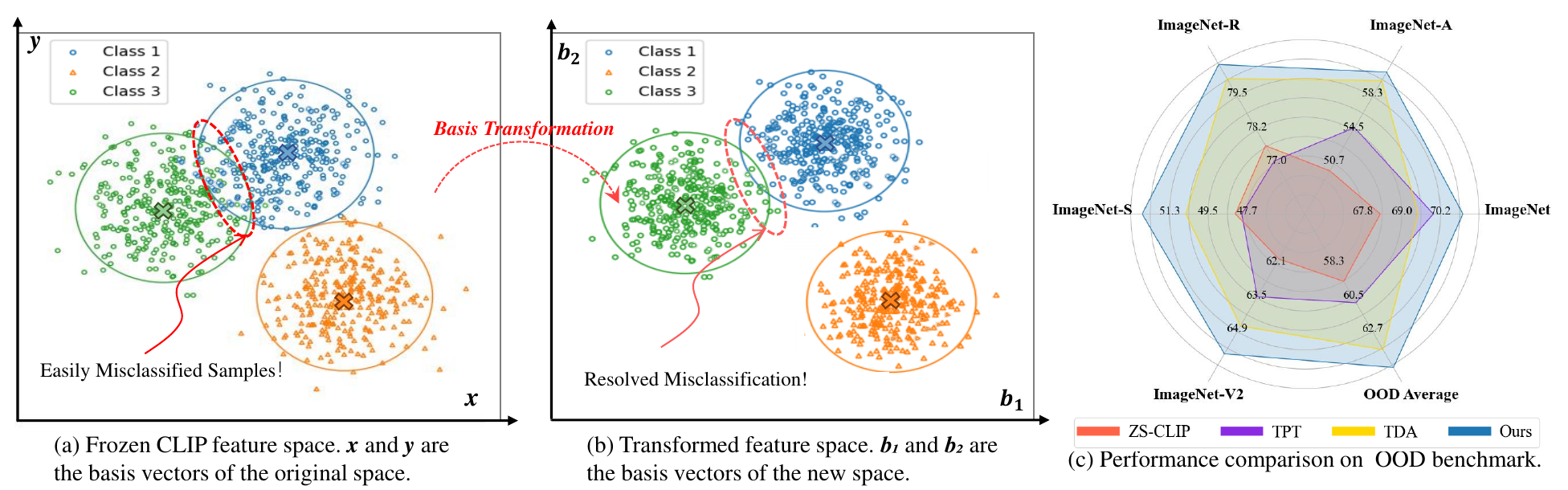}}
\caption{(a) Feature confusion generated in the original CLIP space. It is evident that the original CLIP feature space contains confounding classes. For training-free methods, the lack of capability to adjust the feature space imposes limitations on their subsequent applicability. (b) Feature space reconstructed through transformation. We utilize new basis vectors (such as ${\textit{\textbf{b}}}_{1}$ and ${\textit{\textbf{b}}}_{2}$ in the Fig. (b)) to transform the feature space into a new space. In this space, we can address the confusion present in the original CLIP and overcome the limitations of training-free methods that cannot adjust the feature space.
 (c) Performance comparison on the OOD benchmark. Our method surpasses state-of-the-art methods almost on all datasets.}
\label{fig:head}
\vspace{-3mm}
\end{figure*}

Visual-language models~(VLM), such as CLIP~\cite{clip} and ALIGN~\cite{ALIGN}, have garnered significant attention from researchers due to their strong generalization capabilities in downstream tasks. Various efficient tuning methods, such as prompt tuning~\cite{coop,cocoop,maple} and adapter tuning~\cite{tip,clip_adapter}, have been proposed to leverage training data for enhancing the performance of VLMs on downstream tasks. While those have achieved notable results, their effectiveness is largely limited to the distribution of the current datasets, making it challenging to generalize to domains or distributions beyond the training data. 

In this context, the test-time adaptation~(TTA) was proposed to rapidly adapt to downstream data distributions by utilizing given test samples. Since it requires no training data or annotations, it holds broad application potential in real-world scenarios. The present mainstream TTA methods for VLMs can be divided into two categories: (i) Prompt-tuning TTA paradigm. TPT~\cite{tpt} and DiffTPT~\cite{difftpt} tune prompts through different data augmentation and confidence selection strategies, ensuring consistent predictions across different augmented views of each test data. (ii) Training-free TTA paradigm. TDA~\cite{tda} proposes a training-free dynamic adapter and maintains a high-quality test set to guide the test-time adaptation for VLM. Among them, the prompt-tuning TTA methods~\cite{tpt,difftpt} demand substantial computational resources and time, contradicting the need for rapid adaptation in real-world scenarios. Therefore, this paper focuses on the training-free TTA paradigm. 

Despite its decent performance, the training-free TTA method has a significant drawback, which stems from the characteristics of the training-free paradigm. Due to the inability to perform training, adjusting the feature space becomes very difficult, and thus, the effectiveness of the ``guidance'' entirely depends on the original CLIP feature space. As shown in Fig.~\ref{fig:head}~(a), the test samples inside the red circle are hard for CLIP to predict accurately due to the overlap of decision boundaries. Currently, methods like TDA \cite{tda}, which compare test samples with representative samples in the original feature space to assist prediction, clearly cannot address this inherent drawback.

Inspired by classical machine learning theories~\cite{svm,pca,lle}, we propose a novel training-free test-time adaptation method called \textbf{S}pace r\textbf{O}tation with \textbf{B}asis tr\textbf{A}nsformation~(\textbf{SOBA}). This method utilizes basis transformation techniques~\cite{la} to convert the original nonlinearly separable space into a new linearly separable space, thereby optimizing the decision boundary of the original CLIP model and effectively overcoming the limitations of the training-free TTA paradigm. Specifically, during testing, we first generate one-hot encodings from the CLIP predictions for the test samples. Based on these encodings, we assign pseudo-labels to each sample and store the sample features and their corresponding pseudo-labels. These pseudo-labels and features are then used together to construct a new feature space. To ensure that the reconstructed basis $\mathcal{B}$ better reflects the differences between features of different classes, we perform covariance singular value decomposition on the stored sample set, extract the key information~\cite{pca}, and construct the orthogonal basis $\mathcal{B}$. Based on the basis $\mathcal{B}$, we reconstruct the original feature space, making the features more linearly separable in the new space. Next, we leverage the mean vectors of different classes in the transformed space as class weights to aid classification decisions during testing, thereby enhancing classification accuracy. Additionally, given the sample quality requirements for constructing the orthogonal basis, we maintain a limited dynamic queue to store samples, thereby mitigating the impact of noisy samples on the basis construction process. The queue, guided by an entropy minimization criterion, progressively selects and merges low-entropy prediction samples through enqueue and dequeue operations, thereby improving the quality of pseudo-labels and samples, which in turn enhances the construction of the orthogonal basis. As shown in Fig. \ref{fig:head} (b), the new space formed by the basis $\mathcal{B}$ better highlights the inter-class differences, surpassing the limitations of the original CLIP feature space in Fig. \ref{fig:head} (a) and providing better guidance for the inference process.

In this paper, we present three key contributions. First, we analyze the limitations of current training-free TTA methods in adjusting the feature space. Inspired by machine learning theories, we propose a space rotation method based on basis transformation, which reshapes the feature space and effectively solves the issue of inseparability in the original feature space. Second, our method is efficient. Experiments on the ImageNet dataset show that our method improves testing speed by 13.96\% compared to the SOTA training-free method TDA~\cite{tda}, while its time cost is only 2.15\% of that of the tuning-based method TPT~\cite{tpt}. Finally, our method achieves state-of-the-art (SOTA) performance across various benchmarks (Fig.~\ref{fig:head} (c)), effectively addressing distribution shifts in downstream tasks.

\section{Related Works}
\label{sec:rw}

\noindent \textbf{Vision-Language Model.}
In recent years, vision-language models, as a novel tool capable of processing both visual and linguistic modalities, have garnered widespread attention. These models, such as CLIP~\cite{clip}, ALIGN~\cite{ALIGN}, BLIP~\cite{BLIP}, FILIP~\cite{filip}, etc., leverage self-supervised training on image-text pairs to establish connections between vision and text, enabling the models to comprehend image semantics and their corresponding textual descriptions. This powerful understanding allows vision-language models (e.g., CLIP) to exhibit remarkable generalization capabilities across various downstream tasks~\cite{downsteam1,downsteam2,downsteam3,h2b}. To further enhance the transferability of vision-language models to downstream tasks, prompt tuning and adapter methods have been applied. However, methods based on prompt tuning (such as CoOp~\cite{coop}, CoCoOp~\cite{cocoop}, Maple~\cite{maple}) and adapter-based methods (such as Tip-Adapter~\cite{tip}, CLIP-Adapter~\cite{clip_adapter}) often require large amounts of training data when transferring to downstream tasks, which conflicts with the need for rapid adaptation in real-world applications. Therefore, this paper focuses on test-time adaptation~\cite{tpt}, a method that enables transfer to downstream tasks without relying on training data.

\noindent \textbf{Test-Time Adaptation.}
Test-time adaptation~(TTA) refers to the process by which a model quickly adapts to test data that exhibits distributional shifts~\cite{tta1,memo,ptta,domainadaptor,dota}. Specifically, it requires the model to handle these shifts in downstream tasks without access to training data. TPT~\cite{tpt} optimizes adaptive text prompts using the principle of entropy minimization, ensuring that the model produces consistent predictions for different augmentations of test images generated by AugMix~\cite{augmix}. DiffTPT~\cite{difftpt} builds on TPT by introducing the Stable Diffusion Model~\cite{stable} to create more diverse augmentations and filters these views based on their cosine similarity to the original image. However, both TPT and DiffTPT still rely on backpropagation to optimize text prompts, which limits their ability to meet the need for fast adaptation during test-time. TDA~\cite{tda}, on the other hand, introduces a cache model like Tip-Adapter~\cite{tip} that stores representative test samples. By comparing incoming test samples with those in the cache, TDA refines the model’s predictions without the need for backpropagation, allowing for test-time enhancement. Although TDA has made significant improvements in the TTA task, it still does not fundamentally address the impact of test data distribution shifts on the model and remains within the scope of CLIP's original feature space. We believe that in TTA tasks, instead of making decisions in the original space, it would be more effective to map the features to a different spherical space to achieve a better decision boundary.

\noindent \textbf{Statistical Learning.}
Statistical learning techniques play an important role in dimensionality reduction and feature extraction. Support Vector Machines~(SVM)~\cite{svm} are primarily used for classification tasks but have been adapted for space mapping through their ability to create hyperplanes that separate data in high-dimensional spaces. The kernel trick enables SVM to operate in transformed feature spaces, effectively mapping non-linearly separable data. PCA~\cite{pca} is a linear transformation method that maps high-dimensional data to a new lower-dimensional space through a linear transformation, while preserving as much important information from the original data as possible.

\begin{figure*}[ht]
\centerline{\includegraphics[width=1\linewidth]{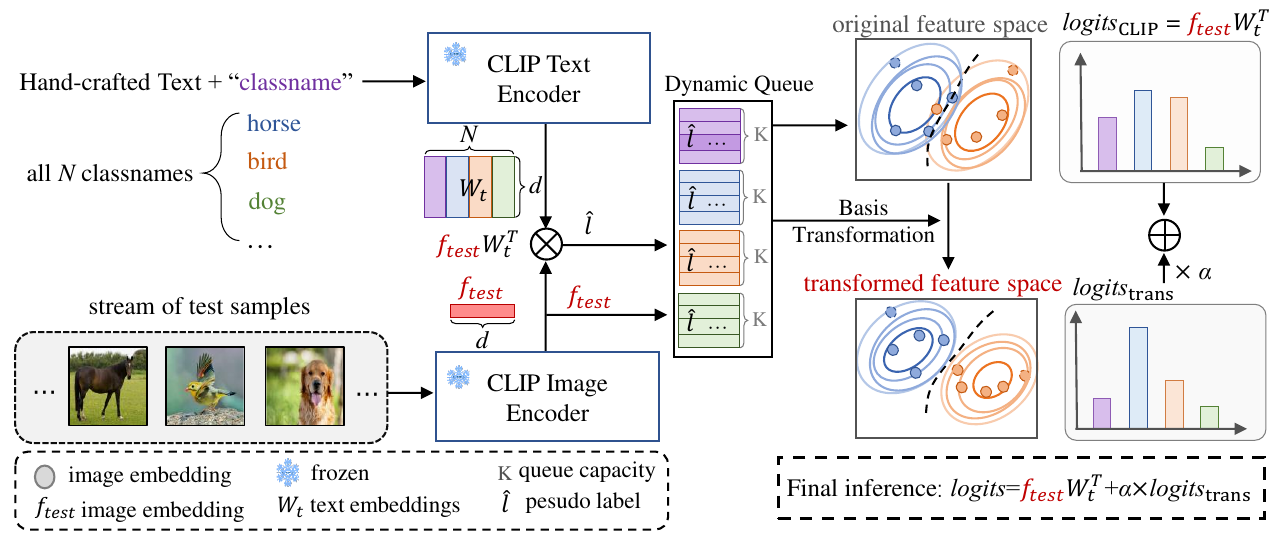}}
\caption{An overview of our method. Our method uses a dynamic queue to store representative samples and generates predictions for test examples based on these samples. This prediction is combined with zero-shot CLIP predictions to produce the final inference. Specifically, we maintain a dynamic queue of representative samples, selected based on minimum entropy of CLIP's predictions. Using these stored samples, we construct a basis transformation to facilitate feature space rotation. As testing progresses, we continuously update and utilize these mappings, allowing the decision boundaries obtained through reconstruction to become more refined and accurate. Finally, we combine the inferences from CLIP with those from the dynamic queue to obtain the final prediction.
}
\label{fig:overview}
\vspace{-4mm}
\end{figure*}

\section{Method}
\label{sec:method}
\subsection{A Training-free Baseline}
\label{sec:Preliminaries}

CLIP~\cite{clip} is a pre-trained vision-language model composed of two parts: a visual encoder and a text encoder, which we represent separately \(E_{v}(\theta_v)\) and \(E_{t}(\theta_t)\). In classification tasks, given a test image $x_{test}$ and \textit{N} classes, CLIP uses \(E_{t}(\theta_t)\) and \(E_{v}(\theta_v)\)  to encode handcrafted text descriptions of the \textit{N} classes and $x_{test}$. After obtaining the corresponding text embeddings ${\mathbf{W}}_{t}$ and visual embedding ${\boldsymbol{f}}_{test}$, CLIP matches the image with the most relevant text description to produce the final prediction as follows:
\begin{equation}\label{eq:clip_logits}
    logits_{\rm{ori}}={\boldsymbol{f}}_{test}{\mathbf{W}}_{t}^{\rm{T}}.
\end{equation}

\indent{Before starting our method, we first construct a training-free baseline method. We utilize a dynamic queue to store representative samples and use these samples to assist in the prediction of test examples. This prediction is combined with the zero-shot CLIP predictions to produce the final inference. Specifically, we dynamically store \textbf{K} test examples for each pseudo-classes, along with their corresponding pseudo-labels $\hat{l}$, using minimum entropy as the criterion. Here, the pseudo-labels are obtained by one-hot encoding the predictions ${\boldsymbol{f}}_{test}{{\mathbf{W}}_{t}}^{\rm{T}}$ for each sample:}
\begin{equation}\label{eq:onehot}
    \hat{l}={\rm{OneHot}}({\boldsymbol{f}}_{test}{{\mathbf{W}}_{t}}^{\rm{T}}).
\end{equation}

When the queue reaches capacity \textbf{K}, we update the queue by replacing the test sample with the highest entropy using the principle of minimizing entropy. Then, during testing, we use an NCM classifier to assist with classification:
\begin{equation}\label{eq:ncm}
    logits_{\rm{NCM}}=\rm{sim}({\boldsymbol{f}}_{test},\mathbf{\mu})
\end{equation}
where $\rm{sim}$ is the cosine similarity, and $\mathbf{\mu}$ is the class mean for each category in the queue.


\subsection{Theoretical Foundation}
During testing, pre-trained models like CLIP often experience reduced generalization due to distribution shifts between downstream tasks and the pre-training dataset. Current approaches focus on improving the selection of augmented views to mitigate this. However, the inference process still faces challenges because the decision boundary remains based on the original CLIP's feature space. For categories with initially poor predictions, the decision boundary in the original feature space limits the effectiveness of augmented view selection, preventing more accurate decisions. This limitation undermines the model's scalability in TTA scenarios.

In this paper, our motivation is to overcome the limitations of the original CLIP feature space for test-time adaptation, aiming to identify a suitable basis. By using the basis to map the original CLIP feature space into a new space, we strive to provide a more effective decision boundary for the inference process. To accomplish this, we propose a training-free feature space rotation method, SOBA, to achieve test-time adaptation of CLIP in downstream tasks.

Before describing our solution, we first present a general explanation of the feature space rotation with basis transformation proposed in this paper. We start by defining a set of feature vectors \( W \in \mathbb{R}^{n \times d} \) as a linear combination of standard orthogonal matrices \({\mathcal{E}}=\{{\mathbf{e}}_{ij}\}_{i,j}\), where \({\mathbf{e}}_{ij} \in \mathbb{R}^{n \times d}\) is defined as a matrix with the \((i,j)\)-th element equal to 1 and all other elements equal to 0. Therefore, we can express \( W \) as:
\begin{equation}\label{eq:ori}
    W= \sum_{i=1}^{n}\sum_{j=1}^{d} w_{ij}{\mathbf{e}}_{ij},
\end{equation}
where, $w_{ij}$ represents the $(i,j)$-th element of $W$, which is also the coefficient of ${\mathbf{e}}_{ij}$.

In this paper, we use an arbitrary basis ${\mathcal{B}}=\{{{\textbf{\textit{b}}}}_{ij}\in {\mathbb{R}}^{n \times d}\}_{i\in[n],j\in[d]}$ to extend $W$. Specifically, ${\mathcal{B}}$ serves as a standard orthogonal basis and must satisfy the following conditions:
\begin{equation}
 \begin{aligned}
    \left \langle{{\textbf{\textit{b}}}},{{\textbf{\textit{b}}}}' \right \rangle=0 \;{\rm{if}}\: {{\textbf{\textit{b}}}} \neq {{\textbf{\textit{b}}}}'\:{\rm{for}}\: {{\textbf{\textit{b}}}},{{\textbf{\textit{b}}}}' \in {\mathcal{B}},\\
    \left \| {{\textbf{\textit{b}}}} \right \| =\sqrt{\left \langle{{\textbf{\textit{b}}}},{{\textbf{\textit{b}}}} \right \rangle}=1 \;{\rm{for}} \,{\rm{all}}\:{{\textbf{\textit{b}}}}\in {\mathcal{B}},
\end{aligned} 
\end{equation}
where, $\left\| \cdot \right\|$ and $\langle \cdot \rangle$ represent the norm and inner product, respectively.

Since the vector hilbert space $\mathcal{H} := \mathbb{R}^{n \times d}$ satisfies the inner product operation $\langle{ \rm{C}}, {\rm{D}} \rangle = \text{trace}({\rm{C}}^{\rm{T}}{\rm{D}})$ (where $ {\rm{C}}, {\rm{D}} \in \mathcal{H}$), we can always express $W \in \mathcal{H}$ as a linear combination of orthogonal matrices in the basis $\mathcal{B}$ under any circumstances. Therefore, Eq.\ref{eq:ori} can be expanded into the following form:
\begin{equation}\label{eq:contrust}
    W=\sum_{{{\textbf{\textit{b}}}}\in{\mathcal{B}}}\left \langle W,{{\textbf{\textit{b}}}} \right \rangle{{\textbf{\textit{b}}}}=\sum_{i=1}^{n}\sum_{j=1}^{d}\left \langle W,{{\textbf{\textit{b}}}}_{ij}\right \rangle{{\textbf{\textit{b}}}}_{ij}.
\end{equation}

We observe that when ${\mathcal{B}} = {\mathcal{E}}$, Eq.\ref{eq:contrust} reduces to Eq.\ref{eq:ori}. Consequently, when all elements in ${\mathcal{B}}$ are orthogonal matrices, we can use ${\mathcal{B}}$ to project $W$ onto a new hypersphere through the mapping $\hat{w}= \{\left \langle W,{{\textbf{\textit{b}}}} \right \rangle \}_{{{\textbf{\textit{b}}}}\in{\mathcal{B}}}$. In Section \ref{sec:SOBA}, we will describe how to use SOBA to address challenges in the TTA task.
\renewcommand\arraystretch{0.9}
\begin{table*}[!t]
  \centering
  \small
  \scalebox{0.9}[0.9]{
  \begin{tabular}{l*{7}c}
    \toprule
    {Method}      & ImageNet   &  ImageNet-A  &  ImageNet-V2  & ImageNet-R 
    & ImageNet-S   
    &\textit{{Average}}  & \textit{{OOD Average}}     \\
  \midrule
  CLIP-ResNet-50 &59.81&	23.24&	52.91	& {60.72}	& 35.48 &	46.43&	43.09           \\ 
  \midrule
  CoOp  &  \textbf{63.33} &  23.06  &   55.40     &  56.60   &  34.67   &     46.61   &     42.43       \\
  
  CoCoOp &  \underline{62.81} &  23.32  &   55.72     &  57.74  &  34.48   &     46.81   &    42.82   \\
Tip-Adapter & 62.03 & 23.13 & 53.97 & 60.35  & 35.74 & 47.04 & 43.30 \\
  \midrule
  TPT  &  60.74 & 26.67  &    54.70   &  59.11   &  35.09   &    47.26    &     43.89    \\
  DiffTPT & 60.80 & \underline{31.06} & \underline{55.80} & 58.80 & 37.10 & 48.71 & 45.69 \\
    \midrule
     $\rm{TDA}^{*}$  & 61.35 & 30.29 & 55.54 & \underline{62.58} & \underline{38.12} & \underline{49.58} & \underline{46.63} \\
    
    \rowcolor{gray!30} ${\mathbf{SOBA~(Ours)}}^{*}$  & 61.85 & \textbf{31.54} & \textbf{55.92} & \textbf{62.91} & \textbf{38.85} & \textbf{50.21} & \textbf{47.31} \\
  \midrule
  \midrule
  CLIP-ViT-B/16 & 68.34&	49.89&	61.88&	{77.65}&	48.24&	61.20&	59.42             \\
  \midrule
  CoOp  &  \textbf{71.51} &  49.71  &   64.20     &  75.21   &  47.99   &   61.72     &   59.28  \\
  
  CoCoOp &  \underline{71.02}  & 50.63  & 64.07       &  76.18   &  48.75   &   62.13     &   59.91  \\
 Tip-Adapter& 70.75 & 51.04 & 63.41 & 77.76 & 48.88 & 62.37 & 60.27 \\
  \midrule
  TPT &  68.98  & 54.77  & 63.45       &  77.06   &  47.94   &   62.44     &   60.81 \\
  DiffTPT  &  70.30  & 55.68  & 65.10       &  75.00   &  46.80   &   62.28     &   60.52 \\					 	 	 	 
    MTA & 69.29 & 57.41 & 63.61 & 76.92 & 48.58 & 63.16 & 61.63 \\
   MTA+Ensemble  & 70.08 & 58.06 & 64.24 & 78.33 & 49.61 & 64.06 & 62.56 \\
   
   \midrule
   $\rm{TDA}^{*}$  & 69.51 & \underline{60.11} & \underline{64.67} & \underline{80.24} & \underline{50.54} & \underline{65.01} & \underline{63.89} \\
 \rowcolor{gray!30} ${\mathbf{SOBA~(Ours)}}^{*}$  & 70.90 & \textbf{61.06} & \textbf{65.83} & \textbf{80.79} & \textbf{52.57} & \textbf{66.23} & \textbf{65.06} \\ 					 	 
    \bottomrule
  \end{tabular}
}
\caption{\textbf{Results on the OOD Benchmark}. Compare the performance of our method with existing methods on OOD benchmark. Our method performs best on both backbones. The best results are in \textbf{bold} and the second-best results are \underline{underlined}. Among the methods we compared, CoOp~\cite{coop} and CoCoOp~\cite{cocoop} are fine-tuned on the training set; TPT~\cite{tpt} and DiffTPT~\cite{difftpt} require backpropagation to update the prompts; TDA~\cite{tda}, and our method do not require any backpropagation to update parameters. \textit{OOD average} refers to the average accuracy on the four OOD datasets from ImageNet, while \textit{average} refers to the average accuracy across all datasets. ``*'' indicates that this method is a training-free approach in test-time adaptation task.
}
\label{tab:ood-main}
\end{table*}
\subsection{Space Rotation with Basis Transformation}
\label{sec:SOBA}
In this section, we first introduce how to construct an appropriate basis vector matrix using SOBA. Then, we explain how to implement it through parameter estimation.


\noindent\textbf{Basis Construction.}
To identify an appropriate basis for reconstructing the matrix \( W \in \mathbb{R}^{n \times d} \), we begin by defining the basis using a pair of unitary matrices. Let \( P \in \mathbb{R}^{n \times n} \) and \( Q \in \mathbb{R}^{d \times d} \) be two arbitrary unitary matrices. We observe that the set \( \mathcal{B} = \{ {\textbf{\textit{b}}}_{ij} := p_{i} q_{j}^{\rm{T}} \in \mathbb{R}^{n \times d} \}_{i \in [n], j \in [d]} \) forms an orthogonal basis, where \( p_{i} \) and \( q_{j} \) represent the \( i \)-th column of \( P \) and the \( j \)-th column of \( Q \), respectively. Consequently, we can express Eq.\ref{eq:contrust} as follows: 

\begin{equation}\label{eq:main}
\begin{split}
W&=\sum_{i=1}^{n}\sum_{j=1}^{d}\left \langle W,{{\textbf{\textit{b}}}}_{ij}\right \rangle{{\textbf{\textit{b}}}}_{ij}\\
&=\sum_{i=1}^{n}\sum_{j=1}^{d}\left \langle W,p_{i} q_{j}^{\rm{T}}\right \rangle p_{i} q_{j}^{\rm{T}}\\
&=\sum_{i=1}^{n}\sum_{j=1}^{d}\hat{w}_{ij}p_{i} q_{j}^{\rm{T}} ,
\end{split}
\end{equation}
where $\hat{w} := \left \langle W, p_{i} q_{j}^{\rm{T}} \right \rangle$. In this case, the basis $\{ p_{i} q_{j}^{\rm{T}} \}_{i,j}$, constructed from a pair of unitary matrices $P$ and $Q$, maps $W$ into the form of $\hat{w}$. Now,  the current challenge is \textit{how to design \( P \) and \( Q \) to achieve a better basis transformation, thereby obtaining an improved space mapping to address distribution shifts in downstream tasks}. 

According to the theory of PCA~\cite{pca}, for a set of feature vectors, we can perform singular value decomposition on their covariance \( C\) to extract the main information:
\begin{equation}\label{eq:pca}
    C=Q_{c}\Sigma Q_{c}^{\rm{T}},
\end{equation}
where \(\Sigma\) is a diagonal matrix with singular values on its diagonal, and \(Q_{c}\) is the corresponding unitary matrix. As observed in the literature~\cite{lowrank}, the features obtained from deep neural networks are often low-rank, meaning that most singular values are close to zero. Due to this low-rank property, for any unitary matrix \( P \), setting \( Q = Q_{c} \) allows us to extract important information from \( W \) under the basis \( \mathcal{B} = \{p_{i} q_{j}^{\rm{T}} \} \) and map this information to \(\hat{w}\). We will introduce how to obtain the covariance matrix \( C \) in Eq.\ref{eq:cov}.

\noindent\textbf{Implementation.}
Subsequently, we will examine the implementation of our proposed method building upon the foundation of the baseline approach in \ref{sec:Preliminaries}. Based on the dynamic queue of the baseline, we utilize SOBA to map the stored features onto a hypersphere, thereby achieving feature reconstruction. The following describes how to implement Eq.\ref{eq:main}.

\renewcommand\arraystretch{0.95}
\begin{table*}[!t]
  \centering
  \resizebox{\linewidth}{!}{
    \begin{tabular}{l*{11}c}
      \toprule
      Method & Aircraft & Caltech101 & Cars & DTD & EuroSAT & Flower102 & Food101 & Pets & SUN397 & UCF101 & \textit{Average} \\
      \midrule
      CLIP-ResNet-50 & 16.11 & 87.26 & 55.89 & 40.37 & 25.79 & 62.77 & 74.82 & 82.97 & 60.85 & 59.48 & 56.63 \\
      \midrule
      CoOp & 15.12 & 86.53 & 55.32 & 37.29 & 26.20 & 61.55 & 75.59 & 87.00 & 58.15 & 59.05 & 56.18 \\
      CoCoOp & 14.61 & 87.38 & 56.22 & 38.53 & 28.73 & 65.57 & 76.20 &  \underline{88.39} & 59.61 & 57.10 & 57.23 \\
      \midrule
      TPT & 17.58 & 87.02 & 58.46 & 40.84 & 28.33 & 62.69 & 74.88 & 84.49 & 61.46 & 60.82 & 57.66 \\
      DiffTPT & 17.60 & 86.89 &  {60.71} & 40.72 & 41.04 & 63.53 & \underline{79.21} & 83.40 &  {62.72} & 62.67 & 59.85 \\
      HisTPT & \textbf{18.10} & 87.20 & \underline
      {61.30} & 41.30 & \textbf{42.50} & 67.60 & \textbf{81.30} & 84.90 & \underline{63.50} & 64.10 & \underline{61.18} \\
      \midrule
       $\rm{TDA}^{*}$  & {17.61} &  \underline{89.70} & 57.78 &  \underline{43.74} & \underline{42.11} & \textbf{68.74} & 77.75 & 86.18 & 62.53 &  \underline{64.18}  &  {61.03}\\
       \rowcolor{gray!30} ${\mathbf{SOBA~(Ours)}}^{*}$ & \underline{17.70} & \textbf{90.18} & \textbf{61.40} & \textbf{44.80} & {41.51}	&  \underline{67.61} &  {77.82} & \textbf{88.69} & \textbf{65.65} & \textbf{66.77} & \textbf{62.20} \\
       
      \midrule
      \midrule
      CLIP-ViT-B/16 & 23.22 & 93.55 & 66.11 & 45.04 & 50.42 & 66.99 & 82.86 & 86.92 & 65.63 & 65.16 & 64.59 \\
      \midrule
      CoOp & 18.47 & 93.70 & 64.51 & 41.92 & 46.39 & 68.71 & 85.30 & 89.14 & 64.15 & 66.55 & 63.88 \\
      CoCoOp & 22.29 & 93.79 & 64.90 & 45.45 & 39.23 & 70.85 & 83.97 &  \underline{90.46} & 66.89 & 68.44 & 64.63 \\
      \midrule
      TPT & 24.78 & 94.16 & 66.87 & \underline{47.75} & 42.44 & 68.98 & 84.67 & 87.79 & 65.50 & 68.04 & 65.10 \\
      DiffTPT &  {25.60} & 92.49 & 67.01 & 47.00 & 43.13 & 70.10 & {87.23} & 88.22 & 65.74 & 62.67 &65.47 \\
      MTA & 25.32 & 94.13 & 68.05	& 45.59	& 38.71	& 68.26	& 84.95	& 88.22	& 64.98	& 68.11	& 64.63 \\
      MTA+Ensemble & 25.20 & 94.21 & {68.47} & 45.90 & 45.36 & 68.06 & 85.00 & 88.24	& 66.60 & 68.69 & 65.58  \\
       HisTPT & \textbf{26.90} & \underline{94.50} & \underline{69.20} & \textbf{48.90} & 49.70 & 71.20 & \textbf{89.30} & 89.10 & 67.20 & 70.10 & \underline{67.61} \\
      \midrule
      $\rm{TDA}^{*}$ & 23.91 &  \underline{94.24} & {67.28} &  {47.40} &  \underline{58.00} &  \underline{71.42} & 86.14 & 88.63 &  \underline{67.62} &  \underline{70.66} &  {67.53} \\
    \rowcolor{gray!30} ${\mathbf{SOBA~(Ours)}}^{*}$ & \underline{25.62} & \textbf{94.60} & \textbf{71.12} & 46.87 & \textbf{59.44} & \textbf{71.66} &  \underline{86.69} & \textbf{92.48}	& \textbf{70.63} &	\textbf{74.12}	& \textbf{69.32} \\
      \bottomrule
    \end{tabular}
  } 
  \caption{\textbf{Results on the Cross-Dataset Benchmark.}
  Compare the performance of our method with existing methods on Cross-Dataset benchmark. Our method achieves the highest average accuracy on both backbones. The best results are in \textbf{bold} and the second-best results are \underline{underlined}. Among the methods we compared, CoOp~\cite{coop} and CoCoOp~\cite{cocoop} are fine-tuned on the training set; TPT~\cite{tpt}, DiffTPT~\cite{difftpt} and HisTPT~\cite{histpt} require backpropagation to update the prompts; TDA~\cite{tda}, and our method do not require any backpropagation to update parameters.  \textit{Average} refers to the average accuracy across all datasets. ``*'' indicates that this method is a training-free approach in test-time adaptation task. 
  }
  \label{tab:cross-dataset}
\end{table*}
\noindent{\textit{Implementation of W}: Similar to the NCM classifier~\cite{ncm}, we use the class mean \(\mathbf{\mu}=\{{\mathbf{\mu}_{k}}\}_{k=1}^{N}\) from the queue as the classifier weights. Setting \( W=\mathbf{\mu} \) in Eq.\ref{eq:main} gives us the mapped class mean \(\hat{\mathbf{\mu}}\). Here, we use the empirical mean to estimate the class mean:}
\begin{equation}\label{eq:means}
    {\mathbf{\mu}_{k}} = \frac{\sum_{i=1}^{M_{k}} \mathbb{I}_{\hat{l}=k} f_{test,i}}{\sum_{i=1}^{M_{k}} \mathbb{I}_{\hat{l}=k}},
\end{equation}
where, $M_{k}$ is the total number of class k. \(\hat{l}\) is the pseudo-label of samples in the queue.

\noindent{\textit{Implementation of \(P=\{p_{i}\}\) and \(Q=\{q_{j}\}\)}:
In practice, we implement Eq.\ref{eq:main} using a very straightforward approach. Due to the properties of the unitary matrix, we can obtain \(PP^{\rm{T}} = I_{n}\) and \(QQ^{\rm{T}} = I_{d}\). Then, we express \( W \) as following:
\begin{equation}\label{eq:main_mat}
    W=PP^{\rm{{T}}}WQQ^{\rm{{T}}}=P{\hat{W}}Q.
\end{equation}

Throughout the process, we set \( P = I_{n} \) and \( Q = Q_{c} \)~(\(Q_{c}\) is obtained from Eq.\ref{eq:pca}). Since \(\hat{w}_{ij}\) is the \((i,j)\)-th element of \(\hat{W}\), we only need to multiply the unitary matrix by \( W \) to achieve the SOBA mapping. During this time, we estimate the covariance matrix using the following approach:
\begin{equation}\label{eq:cov}
    C= \frac{1}{N}\sum_{k=1}^{N} \frac{ {\textstyle \sum_{i=1}^{{M}_{k}}\mathbb{I}_{\hat{l}=k}(f_{test,i}-{\mathbf{\mu}_{k}})(f_{test,i}-{\mathbf{\mu}_{k}})^{\rm{T}}  } }{ {\textstyle \sum_{i=1}^{M_{k}}} \mathbb{I}_{\hat{l}=k}} ,
\end{equation}
where to reduce the computational burden, we adopt the GDA~\cite{gda} assumption for calculating the covariance matrix, which states that all classes follow a distribution with a common covariance.

Ultimately, we obtain the SOBA classifier as follows:
\begin{equation}\label{eq:trans_logits}
    logits_{\rm{trans}}={\rm{Linear}}({\boldsymbol{f}}_{test},\hat{\mathbf{\mu}}).
\end{equation}

Additionally, during the inference process, we update the covariance and mean every 10\% of the test samples to further reduce the computational burden. Ultimately, we employ mixed predictions to consolidate the final logits output. Therefore, the output logits for the test images are calculated as follows:
\begin{equation}\label{eq:last}
    logits= {\boldsymbol{f}}_{test}{{\mathbf{W}}_{t}}^{\rm{T}}+{\alpha}\times logits_{\rm{trans}},
\end{equation}
where \(\alpha\) is a hyperparameter.



\section{Experiment}
\label{sec:experiment}
\begin{table*}[ht]
    \centering
    
    \begin{subtable}{\linewidth}
        \centering
        \small
        \setlength{\tabcolsep}{8pt}
        \scalebox{0.9}[0.9]{
        \begin{tabular}{l*{7}c}
            \toprule
            {Method} & ImageNet & ImageNet-A & ImageNet-V2 & ImageNet-R & ImageNet-S & \textit{Average} & \textit{OOD Average} \\
            \midrule
            Baseline & 69.04 & 60.04 & 64.54 & 80.16 & 49.39 & 64.63 & 63.53 \\
            \textbf{+SOBA~(Ours)} & \textbf{70.90} & \textbf{61.06} & \textbf{65.83} & \textbf{80.79} & \textbf{52.57} & \textbf{66.23} & \textbf{65.06} \\
            \midrule
            \rowcolor{gray!30} Improvement & \textbf{+1.86} & \textbf{+1.02} & \textbf{+1.29} & \textbf{+0.63} & \textbf{+3.18} & \textbf{+1.60} & \textbf{+1.53} \\
            \bottomrule
        \end{tabular}}
        \caption{{Performance improvement of our method over cache baseline on OOD benchmark.}} 
    \end{subtable}
    
    \begin{subtable}{\linewidth}
        \centering
        \small
        \resizebox{\linewidth}{!}{
        \begin{tabular}{l*{11}c}
            \toprule
            Method & Aircraft & Caltech101 & Cars & DTD & EuroSAT & Flower102 & Food101 & Pets & SUN397 & UCF101 & \textit{Average} \\
            \midrule
            Baseline & 24.72 & 94.07 & 67.79 & 45.80 & 55.06 & 71.15 & 86.4 & 88.41 & 67.69 & 70.24 & 67.13 \\
            \textbf{+SOBA~(Ours)} & \textbf{25.62} & \textbf{94.60} & \textbf{71.12} & \textbf{46.87} & \textbf{59.44} & \textbf{71.66} & \textbf{86.69} & \textbf{92.48} & \textbf{70.63} & \textbf{74.12} & \textbf{69.32} \\
            \midrule
            \rowcolor{gray!30} Improvement & \textbf{+0.90} & \textbf{+0.53} & \textbf{+3.33} & \textbf{+1.07} & \textbf{+4.38} & \textbf{+0.51} & \textbf{+0.29} & \textbf{+4.07} & \textbf{+2.94} & \textbf{+3.88} & \textbf{+2.19} \\
            \bottomrule
        \end{tabular}
        }
        \caption{Performance improvement of our method over cache baseline on Cross-Dataset benchmark.} 
    \end{subtable}
    \vspace{-2mm}
    \label{tab:components}
    \caption{\textbf{Performance improvement of our method over cache baseline on both benchmarks.} The experiments employ ViT-B/16 as the backbone. Compared to the baseline, our method exhibits improved performance across all datasets.}
\end{table*}

\subsection{Experimental Setup}
\label{sec:setup}
\noindent{\textbf{Benchmarks.}}
Based on previous work~\cite{tpt,difftpt,tda,mta}, we selected the out-of-distribution (OOD) benchmark and the cross-dataset benchmark as the foundational experiments for our study.
\begin{itemize}
    \item For the \textbf{OOD benchmark}, we tested the effectiveness of our method on out-of-distribution datasets using ImageNet and its four OOD sub-datasets, which include ImageNet-A~\cite{imageneta}, ImageNet-R~\cite{imagenetr}, ImageNet-V2~\cite{imagenetv2}, and ImageNet-S~\cite{imagenetsk}. The purpose of the OOD benchmark is to evaluate the model's generalization ability to data from the same class but different domain distributions.
    \item For the \textbf{cross-dataset benchmark}, we used 10 public datasets to evaluate the cross-dataset classification capability of our method. Each dataset comes from different classes and domains, including: Aircraft~\cite{aircraft}, Caltech101~\cite{caltech101}, Car~\cite{cars}, DTD~\cite{dtd}, EuroSAT~\cite{eurosat}, Flowers102~\cite{flowers}, Food101~\cite{food101}, Pets~\cite{pets}, SUN397~\cite{sun397}, and UCF101~\cite{ucf101}.
\end{itemize}

\noindent{\textbf{Comparison Methods.}}
We compared our method with zero-shot CLIP~\cite{clip}, CoOp~\cite{coop}, CoCoOp~\cite{cocoop}, Tip-Adapter~\cite{tip}, and other state-of-the-art (SOTA) methods in the TTA domain that do not require a training set, including TPT~\cite{tpt}, DiffTPT~\cite{difftpt}, MTA~\cite{mta}, HisTPT~\cite{histpt}, and TDA~\cite{tda}. Among these, Tip-Adapter cannot be evaluated on the cross-dataset benchmark because it is unable to handle unseen classes during the testing phase. Additionally, we do not include MTA in the comparison for experiments with ResNet-50 as the backbone, as there is no data available for MTA on ResNet-50. Furthermore, MTA+Ensemble refers to the ensemble prediction method provided in the MTA paper. Notably, the decision boundary of TDA is based on the original CLIP's feature space, while our method transcends this space.

\noindent{\textbf{Implementation Details}.}
Our method is built upon the pre-trained CLIP~\cite{clip}, where the text encoder of CLIP is a Transformer~\cite{attention_is_all_you_need}, and the image encoder can be either ResNet~\cite{resnet} or Vision Transformer~\cite{vit}. Since our method is training-free, all text prompts are manually crafted. To construct the dynamic queue, we set the batch size to 1. For the OOD benchmark, we conduct a hyperparameter search on ImageNet and apply the resulting hyperparameters to the remaining four OOD datasets. In the case of the cross-dataset benchmark, due to the diversity and complexity of the datasets, the length of the dynamic cache queue varies for each dataset. This will be further explored in an ablation study provided in the Appendix. Additionally, we use top-1 accuracy as the evaluation metric for our experiments, and all experiments are performed on an NVIDIA Quadro RTX 6000 GPU.

\subsection{Comparison with State-of-the-arts}
\label{sec:compare}
We compare our method against zero-shot CLIP, CoOp, CoCoOp, Tip-Adapter, TPT, DiffTPT, MTA, and TDA. Notably, CoOp, CoCoOp, and Tip-Adapter require a training set for optimization, while TPT, DiffTPT, MTA, TDA, and our method do not. Due to methodological constraints, Tip-Adapter cannot be tested on unseen classes, and MTA does not provide accuracy results for the ResNet-50 backbone. Like TPT, DiffTPT, MTA, and TDA, we evaluate our method on both the \textbf{OOD benchmark} and the \textbf{cross-dataset benchmark}.

\noindent \textbf{Results on the Out-of-Distribution Benchmark.}
Table \ref{tab:ood-main} provides a comparison between our method and state-of-the-art (SOTA) approaches across different backbones on ImageNet and four out-of-distribution (OOD) datasets. Our method surpasses existing approaches on all OOD datasets. Notably, it outperforms TDA with an increase of 0.68\% in OOD average accuracy using the ResNet-50 backbone and \textbf{1.17\%} with the ViT-B/16 backbone. Additionally, our approach demonstrates a significant \textbf{3.43\%} improvement over MTA with the ViT-B/16 backbone. These results affirm the effectiveness of exploring new decision boundaries beyond the original CLIP decision surface, validating our approach. 


\begin{table}[t] 
    \small \centering
    \setlength{\tabcolsep}{0.4mm}
    \scalebox{0.9}[0.9]{
    \begin{tabular}{lcccc}
    \toprule
    Method & Training-free & Testing Time & Accuracy & Improved \\
    \midrule
    CLIP-ResNet-50 & \checkmark & \textbf{12min} & 59.81 & 0.  \\
    TPT & \xmark & 12h 50min & 60.74 & 0.93 \\
    DiffTPT & \xmark & 34h 45min & 60.80 & 0.99  \\
    TDA & \checkmark & 16min  &61.35 & 1.54   \\
    \midrule
    \rowcolor{gray!30} \textbf{SOBA~(Ours)} & \checkmark & \underline{13min 46s} & \textbf{61.85} & \textbf{2.04} \\
    \bottomrule
    \end{tabular}}
    \captionof{table}{Comparisons of our method with CLIP-ResNet-50, TPT, DiffTPT and TDA in terms of efficiency and accuracy. The results are achieved with a NVIDIA Quadro RTX 6000 GPU through testing on ImageNet.}
    \label{tab:runtime}
\end{table}

\noindent{\textbf{Efficiency Comparison.}}
As shown in Table \ref{tab:runtime}, to assess the efficiency of our method using ResNet-50 as the backbone, we compared it with three existing test-time adaptation methods on the ImageNet dataset, focusing on inference speed and accuracy. The performance metrics for CLIP-ResNet-50, TPT, DiffTPT, and TDA are sourced from the TDA paper. While our method sacrifices slight efficiency compared to zero-shot CLIP, it achieves a 2.04\% accuracy improvement. Unlike TPT and DiffTPT, which require backpropagation, our method significantly outperforms them in efficiency. Compared to TDA, our method enhances both efficiency and accuracy, improving inference time by 2m 14s and accuracy by 0.5\%. These results demonstrate the efficiency and suitability of our approach for test-time adaptation.

\label{sec:abstudy}

\begin{figure*}[!htbp]
    \centering
    \centerline{\includegraphics[width=1\linewidth]{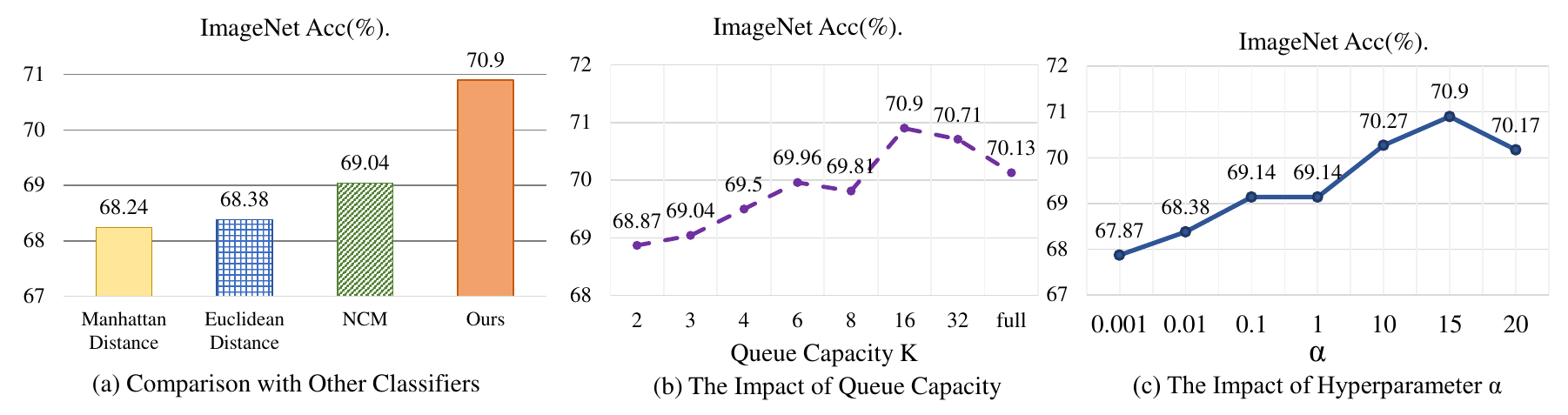}}
    \caption{
    Subfigure (a) shows a comparison with other classifiers, where our SOBA achieves the best performance. Subfigure (b) presents a study on different dynamic queue lengths. Subfigure (c) presents a study on the impact of the hyperparameter $\alpha$. All experiments in the figure are based on ViT-B/16 and conducted on ImageNet~\cite{imagenet}.
}
    \label{fig:ablations_onthers}
    \vspace{-2mm}
\end{figure*}
\noindent \textbf{Results on the Cross-Datasets Benchmark.}
To further validate the feasibility and effectiveness of our approach, we conducted comparisons with SOTA methods across 10 datasets spanning diverse categories and domains. As shown in Table \ref{tab:cross-dataset}, our method consistently outperforms competitors on both backbones tested. Using ResNet-50, our approach achieved top performance on 6 out of 10 datasets, with an average accuracy improvement of \textbf{1.13\%} over TDA. With ViT-B/16, our method led on 7 out of 10 datasets, surpassing TDA with a \textbf{1.79\%} increase in average accuracy. The performance on the cross-dataset benchmark further demonstrates that our method remains effective even when faced with datasets from different classes and domains. Moreover, our method does not require additional training or backpropagation on both benchmarks, making it well-suited for testing adaptation tasks with CLIP.

\subsection{Ablation Studies}
In this section, we conduct ablation experiments to analyze the effectiveness of our design. Our baseline method is the one mentioned in Section ~\ref{sec:Preliminaries}.

\noindent{\textbf{Effectiveness of SOBA.}}
To clearly illustrate the effectiveness of our method, we compare it with a simple yet effective baseline. In Table 3, we report the ablation experiments on the OOD benchmark and cross-dataset benchmark, respectively. Since the baseline method also does not involve backpropagation and is based on the original CLIP feature space, comparing it with this baseline allows us to directly observe the pure benefit of the space rotation provided by SOBA. 

Compared to baseline, our work demonstrates significant improvements across nearly all datasets in both benchmarks. Compared to the baseline, on the OOD benchmark, our two evaluation metrics, \textit{average} and \textit{OOD average}, improved by \textbf{1.6\%} and \textbf{1.53\%}. On the cross-dataset benchmark, we achieved a \textbf{2.19\%} improvement in \textit{average}. Combining our finding with the comparisons to TDA in Section \ref{sec:compare}, that rely on the original CLIP feature space, we can conclude that applying a basis transformation to rotate the original space is a feasible solution to address the TTA problem, and it achieves better performance than the original CLIP feature space.

\noindent{\textbf{Comparison with Other Classifiers.}}
In Fig. \ref{fig:ablations_onthers}(a), we present a comparison of our method with other classifiers. Due to changes in the feature space, directly minimizing the Manhattan~(L1) distance and Euclidean~(L2) distance to class centers is no longer applicable, and it even results in degradation compared to zero-shot CLIP. Our method, compared to the basic NCM classifier, achieves better decision boundaries by utilizing the rotated space, further addressing the test-time adaptation problem.

\noindent{\textbf{Hyperparameter Aensitivity Analysis.}}
\begin{itemize}
    \item \textbf{Queue Capacity K.} In Fig. \ref{fig:ablations_onthers}(b), we report the impact of dynamic queue  Capacity. We find that as the  Capacity of the dynamic queue increases, the overall accuracy shows a trend of first increasing and then decreasing. This can be understood as follows: when the queue  Capacity is small, the stored features are very representative, but as the queue  Capacity increases, some easily confusable features are added, affecting subsequent judgments. In this paper, we select 16 as the storage limit for each class in our dynamic queue on the OOD benchmark.
    \item \textbf{$\alpha$.} In Fig. \ref{fig:ablations_onthers}(c), we illustrate the impact of $\alpha$ from Eq.\ref{eq:last}. Based on the performance on ImageNet, we ultimately select $\alpha=15$ as the final value.
\end{itemize}

\section{Conclusion and Limitation}
\label{sec:conclusion}

In this work, we introduce a space rotation with basis transformation~(SOBA) method, designed to overcome the limitations of the training-free TTA paradigm in the feature space. By leveraging SOBA, we perform a rotation and reconstruction of the original feature space, thereby tackling the adaptation issues that arise from distribution changes during testing. Experimental results across various benchmarks have demonstrated that our method not only outperforms state-of-the-art approaches but is also easy to implement and highly efficient. Nonetheless, our method still requires refinement, as the effectiveness of the space reconstruction relies on pseudo-labels, and resolving this dependency is left for future work.

{
    \small
    \bibliographystyle{ieeenat_fullname}
    \bibliography{main}
}

\clearpage
\setcounter{page}{1}
\maketitlesupplementary

\section{Additional Experimental Details}
\label{sec:addexperimental}
\subsection{Additional Benchmark Details}
\label{sec:addbenchmark}
In this section, we provide detailed information on the two benchmarks used in our work.

\noindent{\textbf{OOD Benchmark.}}
 OOD benchmark is used to validate the model's ability to generalize to data of the same class but with different styles, assessing its robustness and effectiveness against distributional shifts. For the OOD benchmark, we used ImageNet~\cite{imagenet} along with four OOD sub-datasets to evaluate our method's performance on out-of-distribution data. These four datasets include ImageNet-A~\cite{imageneta}, ImageNet-R~\cite{imagenetr}, ImageNet-V2~\cite{imagenetv2}, and ImageNet-S~\cite{imagenetsk}. Below, we provide a brief overview of each OOD dataset.
\begin{itemize}
    \item \textbf{ImageNet-A}~\cite{imageneta}: ImageNet-A is a curated dataset containing 200 challenging classes of images for standard ImageNet-trained models. The dataset is composed of images from the real world that are likely to cause model misclassification, specifically selected to highlight the limitations of traditional models when recognizing out-of-distribution or adversarial samples. 
    \item \textbf{ImageNet-R}~\cite{imagenetr}: ImageNet-R is a dataset derived from ImageNet, specifically designed to test model robustness under significant changes in visual style, covering 200 classes. "R" stands for "Renditions," and the dataset includes images in a variety of artistic styles, such as paintings, cartoons, and sculptures. These images differ significantly from standard ImageNet photographs, making them particularly suitable for evaluating a model's ability to generalize beyond typical photographic representations. 
    \item \textbf{ImageNet-V2}~\cite{imagenetv2}: ImageNet-V2 is a dataset designed to evaluate the consistency and robustness of models trained on the original ImageNet dataset, consisting of 1000 classes. It was created by re-sampling the original ImageNet categories using methods that are similar but not identical to the original collection process. ImageNet-V2 aims to measure the generalization ability of models, as it mimics the distribution of the original dataset while incorporating new, previously unseen samples.
    \item \textbf{ImageNet-S}~\cite{imagenetsk}: ImageNet-S is a dataset derived from ImageNet, containing 1000 classes, specifically designed to evaluate a model's sensitivity to background changes and its ability to focus on salient features. "S" stands for "Sketches," and the dataset consists of black-and-white sketches of the original ImageNet classes. The simplified and abstract nature of the sketches challenges models to classify images based solely on basic contours and shapes, rather than relying on background context or texture information. 
\end{itemize}
\noindent{\textbf{Cross-Dataset Benchmark.}}
The cross-dataset benchmark consists of 10 image classification datasets, each representing a distinct domain and category, designed to evaluate the model's effectiveness and generalization capability across diverse scenarios. The benchmark includes the following datasets: Caltech101 for general image classification; OxfordPets (Pets), StanfordCars (Cars), Flowers102, Food101, and FGVCAircraft (Aircraft) for fine-grained image classification; EuroSAT for satellite imagery classification; UCF101 for action recognition; DTD for texture classification; and SUN397 for scene classification.

For the number of classes and the number of test samples for each dataset in both benchmarks, please refer to the table \ref{tab:supp_datasets}.

\begin{table}[h]
    \centering
    \renewcommand\arraystretch{1.1}{
    \begin{tabular}{l| c c}
    \toprule
    Dataset & Classes & Test Samples\\
    \midrule
    \multicolumn{3}{c}{{OOD benchmark}}\\
    \midrule
    ImageNet & 1,000 & 50,000 \\
    ImageNet-V2 & 1,000  & 10,000\\
    ImageNet-S  & 1,000 & 
    50,000\\
    ImageNet-A & 200  & 7,500 \\
    ImageNet-R & 200 & 30,000 \\
    \midrule
    \multicolumn{3}{c}{{Cross-Dataset benchmark}}\\
    \midrule
    Aircraft & 100 & 3,333 \\
    Caltech101 & 101 & 2,465 \\
    Cars & 196 & 8,041 \\
    DTD & 47 & 1,692 \\
    EuroSAT & 10 & 8,100 \\
    Flowers102 & 102 & 2,463 \\
    Food101 & 101 & 30,300 \\
    Pets & 37 & 3,669 \\
    SUN397 & 397 & 19,850 \\
    UCF101 & 101 & 3,783 \\
    \bottomrule
    \end{tabular}}
    \caption{Datasets Information.}
    \label{tab:supp_datasets}
\end{table}
\subsection{Additional Comparison Methods Details}
\label{sec:addcmparison}
In this section, we provide a detailed description of the methods compared in our work.

\noindent{\textbf{CoOp}~\cite{coop}}: CoOp~\cite{coop} aims to perform automatic prompt optimization for vision-language models (e.g., CLIP) to achieve better few-shot learning and cross-domain generalization. CoOp replaces manually crafted prompt tokens with learnable context vectors while keeping the pre-trained model parameters unchanged. These context vectors are optimized by learning task-specific information from the data, significantly improving model performance.

\noindent{\textbf{CoCoOp}~\cite{cocoop}}: CoCoOp~\cite{cocoop} is an extension of the previous CoOp method. CoCoOp learns a lightweight neural network to generate context prompts conditioned on the input image, making the prompts dynamic rather than static, and adjusting them for each instance. This allows CoCoOp to better adapt to class variations, thereby enhancing the model's generalization ability to new classes.

\noindent{\textbf{Tip-Adapter}~\cite{tip}}: Tip-Adapter~\cite{tip} is designed to adapt the CLIP model for few-shot classification in a training-free manner. Tip-Adapter is based on a key-value cache model, constructing a non-parametric adapter from a small number of training samples without any additional training. It extracts features from few-shot images using CLIP's visual encoder and stores these features along with corresponding pseudo-labels in a cache, leveraging feature retrieval for inference. This approach enables the CLIP model to incorporate few-shot knowledge without retraining, achieving performance comparable to models that require training.

\noindent{\textbf{TPT}~\cite{tpt}}: TPT~\cite{tpt} dynamically adjusts adaptive prompts during testing, using only a single test sample without requiring additional training data or annotations. The method optimizes prompts by minimizing the marginal entropy between augmented views to ensure consistent predictions for different augmented versions of each test sample. Additionally, TPT introduces a confidence selection mechanism to filter out low-confidence augmented samples, thereby reducing the impact of noise.

\noindent{\textbf{DiffPT}~\cite{difftpt}}: DiffTPT~\cite{difftpt} utilizes a pre-trained diffusion model to generate diverse and informative augmented data, while maintaining prediction accuracy through cosine similarity filtering. This method combines traditional data augmentation with diffusion-based augmentation, enabling the model to improve its adaptability when encountering novel data without the need for retraining.

\noindent{\textbf{MTA}~\cite{mta}}: MTA~\cite{mta} employs a robust multimodal MeanShift algorithm to manage augmented views during testing by directly optimizing the quality evaluation of augmented views, referred to as the "inherence score." This method does not require prompt tuning and does not rely on complex training processes, enabling efficient adaptation to new data.

\noindent{\textbf{TDA}~\cite{tda}}: TDA~\cite{tda} uses a lightweight key-value cache to dynamically maintain a small number of pseudo-labels and test sample features. It gradually adapts to test data through progressive pseudo-label refinement, without requiring backpropagation, making it highly efficient. TDA also introduces a negative pseudo-label mechanism, which assigns pseudo-labels to certain negative classes to reduce the impact of noisy pseudo-labels. By combining both positive and negative caches, TDA significantly improves the model's classification accuracy and generalization ability without retraining, while also greatly reducing test time.

\section{Additional Implementation of SOBA}
\label{sec:addsoba}
In this section, we provide a detailed description of the overall process of handling the feature space with basis vectors in our SOBA method.
\subsection{SOBA Process}
\label{sec:soba_process}
The SOBA process includes the following key steps: for each test sample $x_{test}$, the algorithm first extracts the image feature \( f_{test} \) and text features \( W_{t} \) using CLIP's visual encoder \(E_{v}(\theta_v)\) and text encoder \(E_{v}(\theta_v)\), and calculates the original CLIP logits by Eq. \ref{eq:clip_logits}. It then generates pseudo-labels by applying one-hot encoding to the original logits by Eq. \ref{eq:onehot}, and updates the dynamic queue, which stores the image features, pseudo-labels, and logits. After that, we compute the prototype for each pseudo-class and calculates the covariance matrix of the queue by Eq. \ref{eq:means} and Eq. \ref{eq:cov}. 

Next, the prototypes are rotated using the SOBA method to obtain new class prototypes by Eq. \ref{eq:main_mat}, and the transformed logits are computed based on these rotated prototypes by Eq. \ref{eq:trans_logits}. Finally, the algorithm combines the original logits and the transformed logits with a weighting factor $\alpha$ to produce the final prediction. It is worth noting that to ensure the stability and accuracy of the obtained orthogonal basis and class prototypes, we update the prototypes every 10\% of the test samples. This strategy allows the algorithm to optimize the model's adaptability while maintaining computational efficiency, and reduces the impact of bases constructed from too few samples on the final results. The overall process is presented in Algorithm \ref{alg:whole_loop}.
\begin{algorithm*}
\setstretch{1.2}
  \caption{The testing loop of proposed \textbf{SOBA} method for test-time adaptation
    \label{alg:whole_loop}}
  \begin{algorithmic}[1]
   \State \textbf{Input: }{CLIP visual encoder \(E_{v}(\theta_v)\), text encoder \(E_{t}(\theta_t)\), testing dataset $D_{test}$, number of classes $N$, $N$ text descriptions $T$ of $N$ classes, original basis \({\mathcal{E}}\), dynamic queue $L$, hyper-parameter $\alpha$, queue capacity $\rm{K}$.}
  \For{each test sample \( x_{test} \) in \( D_{test} \)}
        \State {Image embedding: } \( f_{test} \) $\gets$ \(E_{v}(\theta_v, x_{test})\)
        \State {Text embeddings: } \( W_{t} \) $\gets$ \(E_{t}(\theta_t, T)\)
        \State {CLIP logits: } \( logits_{ori} \) $\gets$ $f_{test}W_{t}^{\rm{T}}$
        \State {Pseudo-label of \( x_{test} \): } \(\hat{l}\) $\gets$ \texttt{OneHot}($logits_{ori}$)
        \State $L$ $\gets$ \texttt{Update}($L, f_{test}, \hat{l}, logits_{ori}$) \Comment{See Algorithm \ref{alg:update}}
        \For{each pseudo-class \( \hat{l}_{k} \) in \( L \)}
            \State {Get prototype of class \(\hat{l}_{k}\):}
            ${\mathbf{\mu}_{k}}$ $\gets$ $\frac{\sum_{i=1}^{M_{k}} \mathbb{I}_{\hat{l}=k} f_{test,i}}{\sum_{i=1}^{M_{k}} \mathbb{I}_{\hat{l}=k}}$
        \EndFor
        \State {Get covariance $C$ of $L$:}
        $C$ $\gets$ $\frac{1}{N}\sum_{k=1}^{N} \frac{ {\textstyle \sum_{i=1}^{{M}_{k}}\mathbb{I}_{\hat{l}=k}(f_{test,i}-{\mathbf{\mu}_{k}})(f_{test,i}-{\mathbf{\mu}_{k}})^{\rm{T}}  } }{ {\textstyle \sum_{i=1}^{M_{k}}} \mathbb{I}_{\hat{l}=k}} $
        \State Space rotation: $\hat{\mathbf{\mu}}$ $\gets$ \texttt{SOBA}(${\mathbf{\mu}}, C$) \Comment{See Equation \ref{eq:main} and \ref{eq:main_mat}}
        \State {\textbf{SOBA} logits: }\( logits_{trans} \) $\gets$ \texttt{Linear}($f_{test}, \hat{\mathbf{\mu}}$)
        \State {Final inference: } $logits$ $\gets$ $logits_{ori} + {\alpha}\times logits_{\rm{trans}}$
  \EndFor
      \State \Return{$logits$} \Comment{return prediction based on the mode}
  \end{algorithmic}
\end{algorithm*}
\begin{algorithm*}
\setstretch{1.2}
  \caption{Queue update process
    \label{alg:update}}
  \begin{algorithmic}[1]
   \State \textbf{Input: }{CLIP logits of $f_{test}$: $logits_{ori}$, image embedding: $f_{test}$, pseudo-label of $f_{test}$: $\hat{l}$, old queue: $L^{t-1}$, queue capacity: $\rm{K}$.}
    \If{$|L^{t-1}_{\hat{l}}| < K$}
    \State $L^{t}_{\hat{l}}$ $\gets$ \texttt{EnQueue}($f_{test}$, $L^{t-1}_{\hat{l}}$)
    \Else
        \State ${\rm{H}}_{max}$ $\gets$ \texttt{max}(H($L^{t-1}_{\hat{l}}$)) \Comment{Get the maximum entropy in $L^{t-1}_{\hat{l}}$.}
        \If{${\rm{H}}(logits_{ori})<{\rm{H}}_{max}$}
          \State {Dequeue feature with ${\rm{H}}_{max}$: } 
          $L^{t-1}_{\hat{l}}$ $\gets$ \texttt{DeQueue}($f_{test}^{ent}$, $L^{t-1}_{\hat{l}}$)
          \State {Enqueue feature $f_{test}$: }$L^{t}_{\hat{l}}$ $\gets$ \texttt{EnQueue}($f_{test}$, $L^{t-1}_{\hat{l}}$)
        \Else
            \State $L^{t}_{\hat{l}}$ $\gets$ $L^{t-1}_{\hat{l}}$
        \EndIf
    \EndIf
    \State \Return{$L^{t}$} \Comment{update the queue}
  \end{algorithmic}
\end{algorithm*}
\subsection{Queue Update Process}
\label{sec:queue_update}
In this section, we explain how to perform enqueue and dequeue operations on the queue.

First, for each test feature $x_{test}$, the algorithm checks whether the queue \( L^{t-1}_{\hat{l}} \) corresponding to the current pseudo-label \( \hat{l} \) is full. If the queue is not full, the current feature \( f_{test} \) and its corresponding pseudo-label \( \hat{l} \) are simply enqueued, generating a new queue \( L^{t} \). If the queue is full, the algorithm first calculates the maximum entropy \( H_{max} \) in the queue, which represents the average uncertainty of the current features. Then, the algorithm compares the entropy of the current feature's logits \( H(logits_{ori}) \) with the maximum entropy \( H_{max} \). If the current feature's entropy is smaller than the maximum entropy, it indicates that the feature is more certain, and the algorithm removes the feature with the highest entropy from the queue and enqueues the current feature; otherwise, the queue remains unchanged. Finally, the algorithm returns the updated queue \( L^{t} \), which helps manage the updates of features and pseudo-labels, ensuring that the queue adapts to new data over time. The overall process is presented in Algorithm \ref{alg:update}.

\section{Additional Ablation Study}
\label{sec:abbstudy}

\renewcommand\arraystretch{0.95}
\begin{table*}[t]
  \centering
  \resizebox{\linewidth}{!}{
    \begin{tabular}{l*{11}c}
      \toprule
      K & Aircraft & Caltech101 & Cars & DTD & EuroSAT & Flower102 & Food101 & Pets & SUN397 & UCF101 & \textit{Average} \\
      \midrule
        2 & 24.72 &	93.59 &	67.79 &	45.80 &	55.06 &	71.34 &	86.40 &	91.61 &	67.79 &	73.09 &	67.72 \\
        4 & 24.99 &	93.91 &	68.90 &	45.33 &	54.30 &	71.50 &	86.59 &	91.63 &	68.15 &	72.40 &	 67.77 \\
        6 & 25.08 &	94.24 &	70.26 &	45.39 &	54.63 &	71.54 &	86.69 &	91.28 &	69.30 &	72.93 & 68.13 \\
        8 & 25.32 &	\textbf{94.60} &	70.17 &	45.98 &	58.79 &	\textbf{71.68} &	86.57 &	91.77 &	69.41 &	73.83 &	68.81 \\
        16 & \textbf{25.62} & \textbf{94.60} & 71.12 & \textbf{46.87} & \textbf{59.44} & 71.66 & 86.69 & 92.48 & \textbf{70.63} &	\textbf{74.12} &	\textbf{69.32} \\
        32 & 25.27 & 93.31 & \textbf{71.22} & 46.34 & 58.28 & 71.38 & \textbf{86.79} & \textbf{92.80} & 69.35 &	72.77 &	68.75 \\
        full & 25.21 &	93.31 &	70.64 &	45.81 &	58.11 &	71.38 &	86.53 &	92.74 &	68.91 &	72.55 &	68.52 \\
      \bottomrule
    \end{tabular}
  } 
  \caption{\textbf{Results on the Cross-Dataset Benchmark.} The performance of SOBA with different K on the Cross-Dataset benchmark. Due to the complexity of the datasets in the cross-dataset benchmark, the performance of each dataset may vary differently as the queue capacity increases. The backbone used in the experiments is ViT-B.
  }
  \label{tab:supp_K_of_cd}
\end{table*}
This section supplements the ablation experiment on queue capacity on the cross-dataset benchmark. Due to the complexity of the datasets in the cross-dataset benchmark, the performance of each dataset may vary differently as the queue capacity increases. For the Pets dataset~\cite{pets}, the best accuracy is achieved when the queue capacity per class is 32. We believe the reason is that the differences between different classes in the Pets dataset are significant, as these classes not only exhibit distinct visual features (such as fur color, shape, and body size), but also show considerable diversity in terms of image background, posture, and camera angle. Therefore, increasing the queue capacity can better capture the information of the feature space, allowing the reconstructed basis and class prototypes to more effectively reflect the differences between classes. Finally, we used $K=16$ as the overall queue capacity for the cross-dataset benchmark.

%

\end{document}